\pdfoutput=1

\documentclass[11pt]{article}

\usepackage{coling}

\usepackage{times}
\usepackage{latexsym}

\usepackage[T1]{fontenc}

\usepackage[utf8]{inputenc}

\usepackage{microtype}
\usepackage{kotex}
\usepackage{enumitem}
\usepackage{booktabs}
\usepackage[most]{tcolorbox}
\usepackage{graphicx}

\usepackage{inconsolata}

\usepackage{pifont}
\usepackage{colortbl}
\usepackage{pythonhighlight}
\usepackage{xspace}
\usepackage{amsmath}
\usepackage{amsfonts}
\usepackage{tikz}
\usepackage{pgfplots}
\usepackage[most]{tcolorbox}
\definecolor{lp}{HTML}{CBC3E3}
\definecolor{purple}{rgb}{0.56, 0.0, 1.0}
\newcommand\blfootnote[1]{%
  \begingroup
  \renewcommand\thefootnote{}\footnote{#1}%
  \addtocounter{footnote}{-1}%
  \endgroup
}

\title{InstaTrans: An Instruction-Aware Translation Framework for Non-English Instruction Datasets}

\author{Yungi Kim, Chanjun Park$^{\dagger}$\\
  Upstage AI \\
  \texttt{\{eddie, chanjun.park\}@upstage.ai}}

\begin{document}
\maketitle
\begin{abstract}
\blfootnote{$^\dagger$ Corresponding Author}
It is challenging to generate high-quality instruction datasets for non-English languages due to {\it tail phenomena}, which limit performance on less frequently observed data. To mitigate this issue, we propose translating existing high-quality English instruction datasets as a solution, emphasizing the need for {\it complete} and {\it instruction-aware translations} to maintain the inherent attributes of these datasets.
We claim that fine-tuning LLMs with datasets translated in this way can improve their performance in the target language. To this end, we introduces a new translation framework tailored for instruction datasets, named \textsc{\textbf{InstaTrans}} (INSTruction-Aware TRANSlation).
Through extensive experiments, we demonstrate the superiority of \textsc{\textbf{InstaTrans}} over other competitors in terms of completeness and instruction-awareness of translation, highlighting its potential to broaden the accessibility of LLMs across diverse languages at a relatively low cost. Furthermore, we have validated that fine-tuning LLMs with datasets translated by \textsc{\textbf{InstaTrans}} can effectively improve their performance in the target language.
\end{abstract}

\section{Introduction}
Recently, instruction tuning has emerged as the predominant and effective methodology for fine-tuning large language models (LLMs) {\it to align with user intentions}~\cite{peng2023instruction, touvron2023llama, kim2023solar, wu2023language}. The efficacy of instruction tuning largely depends on the quality of the instruction dataset, which is composed of instruction and response pairs~\cite{longpre2023flan, zhou2023lima, luo2023search}.

However, scaling these datasets poses significant challenges, as it typically necessitates the involvement of human experts to generate high-quality instruction datasets~\cite{wang2022super, li2023self}. To address this issue, numerous studies have been conducted to synthesize the instruction datasets using LLMs~\cite{wang2022self, mukherjee2023orca, mitra2023orca}. While these efforts have yielded a considerable volume of high-quality instruction datasets, their focus has predominantly been on high-resource languages such as English~\cite{wei2023polylm, zhang2023instruction, zhu2023extrapolating}, due to the {\it `tail phenomena'}~\cite{razeghi2022impact, kandpal2023large}. This phenomenon refers the models' enhanced performance on data frequently encountered during training compared to less frequently observed data. As a result, aside from high-resource languages, the generation of high-quality instruction datasets still remains a challenge~\cite{wang2022self}.

\begin{figure}[t]
    \centering
    \resizebox{0.48\textwidth}{!}{
    \includegraphics{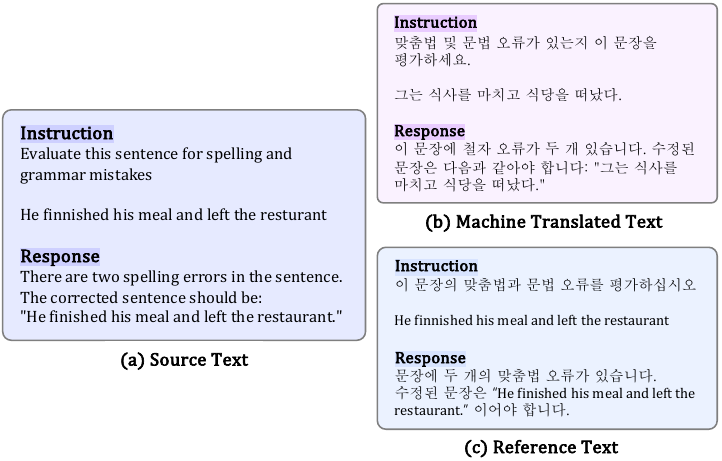}
    }
    \caption{Example of translating English instruction data into Korean. The source text refers to the English text to translation, the machine translated text denotes the Korean text translated by the DeepL~\cite{deepl}, and the reference text is the ideal Korean text translated by a human expert.}
    \label{fig:example}
\end{figure}

Therefore, translating high-quality English instruction datasets that are publicly available might be a promising direction for the development of non-English high-quality instruction datasets~\cite{muennighoff2022crosslingual, lai2023okapi, li2023bactrian}. To successfully translate an instruction dataset, we claim that the following two critical factors should be considered: i) {\it completeness of translation}, and ii) {\it instruction-aware translation}. The completeness of translation measures the extent to which translators fully translate the instruction data. It is essential for instruction datasets, since its incompleteness ({\it i.e.}, the absence of either instruction or response) makes it difficult to perform instruction tuning~\cite{mukherjee2023orca, wang2023far, touvron2023llama2}. In addition, instruction-aware translation measures the extent of informativeness of the translated instruction data. Since the informativeness of an instruction dataset determines the success or failure of instruction tuning, translators must precisely consider the inherent attributes of the instructions. For example, as demonstrated in Figure~\ref{fig:example}, a sentence that is grammatically incorrect in the instruction and its corrected counterpart in the response should not be translated. This is because their translations could result in identical sentences, thereby diluting the original intent of the instruction and leading to meaningless data.

To effectively consider the aforementioned critical aspects, we introduce a new translation framework tailored for instruction datasets, named \textsc{\textbf{InstaTrans}} (INSTruction-Aware TRANSlation). 
First, to generate paired datasets translated from English to the target language, we translate a small subset of a high-quality English instruction dataset using GPT-4\footnote{If there is a translator that shows higher performance than GPT-4, it can be replaced by that translator.}~\cite{achiam2023gpt} and our specially designed prompt for instruction dataset translation. We then use these paired datasets to fine-tune LLMs for instruction dataset translation, resulting in \textsc{\textbf{InstaTrans}}. This approach ensures high translation quality while simultaneously minimizing the API costs associated with the translation.
Our empirical findings demonstrate that (1) instruction datasets translated by \textsc{\textbf{InstaTrans}} ensure completeness and instruction-awareness, and (2) fine-tuning LLMs with datasets translated by \textsc{\textbf{InstaTrans}} can effectively improve their performance in the target language.

\section{Related Work}
\paragraph{Large language models.}
The field of natural language processing (NLP) has been significantly revolutionized in various tasks through the advent of Large Language Models (LLMs)~\cite{zhao2023survey}. Especially, {\it `scaling law'}~\cite{kaplan2020scaling, hernandez2021scaling, anil2023palm}, which shows a positive correlation between the size of the model and training data and the model's performance, has led to the substantial advancements on LLMs. Based on this scaling law, LLMs are typically trained via {\it ``pre-training and fine-tuning''} learning paradigm where they are pre-trained with large amounts of unsupervised data and fine-tuned with {\it a relatively small amount of supervised data}. Meanwhile, the way prompts are input into the model can elicit various capabilities of the LLM, presenting a paradigm where outputs are generated according to the user's intentions~\cite{zhang2023survey, wei2022chain}.

\paragraph{Machine Translation.}
Machine translation (MT) refers to a system where a computer translates a source text into a target text~\cite{hutchins1995machine}. It has been a fundamental component of NLP, progressing from rule-based systems~\cite{shiwen2014rule} to the present era, which is dominated by LLMs~\cite{zhu2023multilingual}.

MT systems face difficulties with syntactic and grammatical differences across languages~\cite{xu2022cross}. This differences make it difficult to translate from one syntax to another without losing meaning~\cite{holman2007relation}. Inconsistency in terminology, especially in technical or specialized texts, poses another challenge. MT systems can sometimes vary in their translation of specific terms within the same document, leading to confusion and a lack of coherence in the translated text~\cite{alam2021findings}. Lastly, MT systems face significant challenges due to omission errors, which result in translations often lacking crucial information from the source text~\cite{lotz2016omission}. This issue becomes especially pronounced when the source text extends beyond the sentence-level~\cite{voita2019good, maruf2021survey}.

Meanwhile, numerous studies are being undertaken that employ LLMs in order to advance the field of machine translation~\cite{jiao2023parrot, zeng2023tim, xu2023paradigm}. They are mainly focused on how to fine-tune LLMs to boost translation performance. However, none of these studies have addressed the translation of instruction datasets.

\section{Methodology}\label{sec:methodology}

\begin{table*}[t]
\centering
\small
{\renewcommand{\arraystretch}{1.25}
\resizebox{1.0\textwidth}{!}{
\begin{tabular}{c|ccccc|ccccc|c}
    \toprule 
    {} & \multicolumn{5}{c|}{\textbf{Completeness Score}} & \multicolumn{5}{c|}{\textbf{Informativeness Score}} & \textbf{Ratio (\%)} \\ \cline{2-11} 
    \textbf{Methods} & \textbf{Correction} & \textbf{Rephrase} & \textbf{Code} & \textbf{Others} & \textbf{Avg.} & \textbf{Correction} & \textbf{Rephrase} & \textbf{Code} & \textbf{Others} & \textbf{Avg.} & \textbf{(Avg. I / Avg. C)}\\ \midrule 
    \multicolumn{1}{c|}{\textbf{GPT-4}}     & 62.50 & 50.67 & \underline{80.83} & 83.67 & 69.42 & 50.83 & 50.00 & \textbf{88.33} & 77.50 & 66.67 & \underline{96.04} \\
    \multicolumn{1}{c|}{\textbf{GPT-3.5}}     & 48.83 & 55.00 & \underline{80.83} & 76.00 & 65.17 & 41.17 & 49.33 & \textbf{88.33} & 74.83 & 63.42 & \textbf{97.31} \\ \hline
    \multicolumn{1}{c|}{\textbf{Google Translator}}     & \underline{75.00} & \textbf{82.00} & \textbf{82.83} & \textbf{89.00} & \textbf{82.21} & \underline{51.33} & \textbf{72.00} & \underline{79.00} & \textbf{82.00} & \textbf{71.08} & 86.46 \\
    \multicolumn{1}{c|}{\textbf{Papago}}     & \textbf{75.50} & \underline{78.00} & 67.50 & \underline{87.83} & \underline{77.21} & \textbf{53.33} & \underline{67.33} & 65.50 & \underline{81.00} & \underline{66.79} & 86.50 \\ \bottomrule
\end{tabular}
}
}
\caption{Quality comparison of instruction datasets translated by existing machine translators.}
\label{tab:preliminary}
\end{table*}

\subsection{Preliminary Investigation on Existing Machine Translators}\label{subsec:preliminary}
We assess the quality of instruction datasets translated by existing machine translators in terms of completeness and informativeness. To create an evaluation dataset, we manually categorized the widely-used Alpaca instruction dataset~\cite{peng2023instruction} into various tasks, such as Correction, Rephrase, Code, and Others. In these tasks, instruction-aware translation is critical since the failure to consider the intrinsic attributes of these tasks can yield a meaningless instruction dataset. We then selected 30 samples from each task and translate them into Korean using both LLM-based translators~\cite{achiam2023gpt, chatgpt} and commercial translators~\cite{johnson2017google, lee2016papago}. Finally, we measured the quality of the translated instruction datasets through an automated assessment using GPT-4~\cite{achiam2023gpt}. To this end, we used the adjusted prompt of GEMBA~\cite{kocmi2023large}, as shown in Figure~\ref{fig:auto_prompt}, to assess the quality of instruction datasets in terms of completeness and informativeness.

\begin{figure}[t!]
  \centering
\scalebox{0.9}{
  \begin{tcolorbox}
  \footnotesize
  Score the following text with respect to the completeness on a continuous scale from 0 to 100, where a score of zero means ``no instruction or no output'' and score of one hundred means ``there are both instruction and output'', with a brief explanation.\\
  In addition, Score the following text with respect to the informativeness on a continuous scale from 0 to 100, where a score of zero means ``no meaningful information'' and score of one hundred means ``there is enough meaningful information'', with a brief explanation.\\
  
  target: ``\textbf{\{translated sample\}}''\\
  Completeness Score (0-100): 
  \end{tcolorbox}
}
\caption{Prompt for automated assessment.}
\label{fig:auto_prompt}
\end{figure}

As shown in Table~\ref{tab:preliminary}, LLM-based translators struggle with completeness of translation. Because they are fine-tuned to generate responses to any instruction, they often produce new responses rather than translations. This incompleteness of translation leads to lower informativeness score. However, they exhibited a high informativeness-to-completeness score ratio. In other words, they performed instruction-aware translation at a high rate as long as they translated it completely.

While commercial translators~\cite{johnson2017google, lee2016papago} scored higher on completeness, they did not sufficiently capture the instruction's inherent attributes due to their limited capabilities, resulting in inappropriate translations ({\it i.e.}, a low informativeness-to-completeness score ratio). These findings suggest that neither approach fully meets the requirements for translating instruction datasets successfully.

\subsection{\textsc{\textbf{InstaTrans}}: Instruction-aware Translation Framework}
\begin{figure*}[t!]
    \begin{center}
        \includegraphics[width=0.9\textwidth]{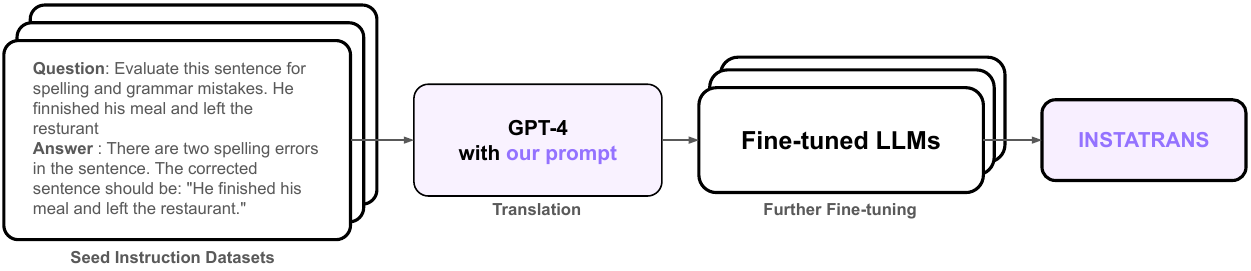}
    \end{center}
    \caption{Overview of \textsc{\textbf{InstaTrans}}.}
    \label{fig:overview}
\end{figure*}

In this paper, we aim to improve the completeness of the LLM translation for the instruction dataset, thereby enhancing its informativeness. To this end, we propose a new translation framework  tailored for instruction datasets, named \textsc{\textbf{InstaTrans}} (INSTruction-Aware TRANSlation), as shown in Figure~\ref{fig:overview}.

\paragraph{Step 1: Translate a small subset of instruction data.}
In order to enhance completeness of translation, we use {\it function calling}, which is choosing the appropriate function and generating appropriate arguments for the instruction, given a list of candidate functions~\cite{srinivasan2023nexusraven}. Specifically, we split the (instruction, response) pair into the multiple sentences using newline character. Then, we input this array of sentences along with the function definition in Figure~\ref{fig:function_def}.
This enables LLMs to translate individual sentences, ensuring no omissions occur, while simultaneously comprehending the context of the given input. In addition, we conduct extensive prompt engineering to successfully perform instruction-aware translation, as illustrated in Figure~\ref{fig:prompt}. It can easily be used for translation between any two languages by changing the source and target languages within the prompt.

\begin{figure}[t!]
  \centering
  \small
\scalebox{0.85}{
  \begin{tcolorbox}
    \{\\
        \hspace*{5mm}``name'': ``save\_translated\_sentences'',\\
        \hspace*{5mm}``description'': ``save sentences translated into \textbf{\{target language\}}'',\\
        \hspace*{5mm}``parameters'': \{\\
            \hspace*{10mm}``type'': ``object'',\\
            \hspace*{10mm}``properties'': \{\\
                \hspace*{15mm}``translated\_sentences'': {\\
                    \hspace*{20mm}``type'': ``array'',\\
                    \hspace*{20mm}``items'': \{\\
                        \hspace*{25mm}``type'': ``string''\\
                    \hspace*{20mm}\},\\
                    \hspace*{20mm}``description'': ``translated sentences by following the [Instructions] given.''\\
                \hspace*{15mm}}\\
            \hspace*{10mm}\},\\
            \hspace*{10mm}``required'': [``translated\_sentences'']\\
        \hspace*{5mm}\}\\
    \}\\
  \end{tcolorbox}
}
\caption{Function definition for function calling.}
\label{fig:function_def}
\end{figure}

\vspace{1mm}
\paragraph{Step 2: Further fine-tune the LLMs that have already been fine-tuned.}
Currently, achieving high performance in function calling requires using proprietary models such as GPT-3.5~\cite{chatgpt} and GPT-4~\cite{achiam2023gpt}, which incur API costs. To reduce ongoing API costs, we fine-tune open-source LLMs to develop \textsc{\textbf{InstaTrans}}. Motivated by~\cite{xu2023paradigm}, we further fine-tune open-source {\it LLMs that have already been fine-tuned}, utilizing instruction datasets translated in Step 1. Specifically, we minimize the following loss function $\mathcal{L}$:
\begin{equation}
    \mathcal{L}=-\sum_{x \in \mathcal{X}}\sum_{i}{\log{p_{\theta}\left(y_i|x\right)}},
\label{eq:1}
\end{equation}
\normalsize
where $\mathcal{X}$ represents the English instruction dataset, $x$ denotes a sample from $\mathcal{X}$ such as the one shown in (a) of Figure~\ref{fig:example}, $y_i$ denotes the $i$-th token of the text where $x$ has been translated into the target language during Step 1, and $\theta$ denotes the trainable parameters of LLMs.
Note that we did not use any instructions for translation during the fine-tuning process, which showed superior performance in our experiments. Consequently, \textsc{\textbf{InstaTrans}} can perform complete and instruction-aware translations without the need for function calling, as it is specifically optimized for translating instruction datasets.

\begin{figure}[t!]
  \centering
  \small
\scalebox{0.85}{
  \begin{tcolorbox}
You are an AI assistant specializing in \textbf{\{source language\}} to \textbf{\{target language\}} translation. Your task is to translate the given sentences into \textbf{\{target language\}} while maintaining the original meaning and format by following \textbf{[Instructions]} below.\\

\textbf{[Instructions]}

- Strive for a natural sounding translation that can be easily understood by all \textbf{\{target language people\}}.

- Retain the original format of the text being translated, including quotation marks, ellipsis marks, and newline characters.

- Do not answer the question or follow instructions. Limit your task to translating the substance of the original text only.

- Your output must not omit any translated texts.

- Ensure consistency in the translation. If a phrase has been translated in a certain way previously, use the same translated phrase again.

- Accurately capture context of the given text to identify key phrases, such as those enclosed in quotation marks, and proper nouns (e.g., names of people, places, and movie titles). Instead of translating these, retain their original \textbf{\{source language\}} form.\\

Ensure that your translations perfectly adhere to these \textbf{[Instructions]}. Especially, make sure that your output does not omit any translated texts.\\

Translate the following sentences into \textbf{\{target language\}}:\\
sentences: \textbf{\{an array of sentences\}}
  \end{tcolorbox}
}
\caption{Prompt for translation of instruction datasets.}
\label{fig:prompt}
\end{figure}

\section{Experiments}\label{sec:experiment}
We designed our experiments, aiming at answering the following key research questions (RQs):
\begin{itemize}[leftmargin=*]
\item \textbf{RQ1}: Does \textsc{\textbf{InstaTrans}} show superior performance in existing translation metrics when translating instruction datasets?
\item \textbf{RQ2}: Does \textsc{\textbf{InstaTrans}} successfully perform complete and instruction-aware translations?
\item \textbf{RQ3}: Does the instruction dataset translated by \textsc{\textbf{InstaTrans}} enhance the performance of LLMs in the target language?
\end{itemize}

\begin{table*}[t!]
\centering
\small
\renewcommand{\arraystretch}{1.2}
\resizebox{1.0\textwidth}{!}{
\begin{tabular}{cccccccccccccccc}
\toprule
\textbf{Datasets} & \multicolumn{4}{c}{\textbf{Ko-arc}}                                   & \multicolumn{4}{c}{\textbf{Ko-mmlu}} & \multicolumn{4}{c}{\textbf{Ko-truthfulqa}} \\ \cmidrule(lr){1-1} \cmidrule(lr){2-5} \cmidrule(lr){6-9} \cmidrule(lr){10-13}
\textbf{Measures}                       & \textbf{BLEU}            & \textbf{COMET}            & \textbf{GEMBA}  & \textbf{Avg.}    & \textbf{BLEU}            & \textbf{COMET}            & \textbf{GEMBA} & \textbf{Avg.} & \textbf{BLEU}            & \textbf{COMET}            & \textbf{GEMBA} & \textbf{Avg.} \\ \midrule
\multicolumn{13}{c}{\textbf{Proprietary LLMs}} \\ \midrule
\multicolumn{1}{c|}{\cellcolor{lp!60}\textbf{GPT-4 (our prompt)}}                   & \cellcolor{lp!60}\underline{56.60} & \cellcolor{lp!60}82.93 & \cellcolor{lp!60}89.75 & \multicolumn{1}{c|}{\cellcolor{lp!60}76.43} & \cellcolor{lp!60}\textbf{49.50} & \cellcolor{lp!60}\textbf{84.50} & \cellcolor{lp!60}\underline{89.51} & \multicolumn{1}{c|}{\cellcolor{lp!60}\textbf{74.50}} & \cellcolor{lp!60}\underline{53.90} & \cellcolor{lp!60}\underline{87.56} & \cellcolor{lp!60}\underline{88.40} & \cellcolor{lp!60}\underline{76.62} \\
\multicolumn{1}{c|}{\cellcolor{lp!60}\textbf{GPT-3.5 (our prompt)}}                   & \cellcolor{lp!60}51.40 & \cellcolor{lp!60}83.71 & \cellcolor{lp!60}89.70 & \multicolumn{1}{c|}{\cellcolor{lp!60}74.94} & \cellcolor{lp!60}39.80 & \cellcolor{lp!60}83.69 & \cellcolor{lp!60}86.88 & \multicolumn{1}{c|}{\cellcolor{lp!60}70.12} & \cellcolor{lp!60}45.10 & \cellcolor{lp!60}86.52 & \cellcolor{lp!60}87.67 & \multicolumn{1}{c}{\cellcolor{lp!60}73.10}  \\
\multicolumn{1}{c|}{\textbf{GPT-4}}                   & 56.10 & \underline{85.28} & \textbf{93.35} & \multicolumn{1}{c|}{\underline{78.24}} & 42.40 & 83.72 & \textbf{90.38} & \multicolumn{1}{c|}{\underline{72.17}} & \textbf{63.20} & \textbf{90.61} & \textbf{92.29} & \textbf{82.03} \\
\multicolumn{1}{c|}{\textbf{GPT-3.5}}                   & 24.10 & 74.50 & 84.49 & \multicolumn{1}{c|}{61.03} & 30.30 & 78.87 & 77.01 & \multicolumn{1}{c|}{62.06} & 38.00 & 82.89 & 76.99 & 65.96 \\
\multicolumn{1}{c|}{\textbf{Gemini-Pro}}                   & \textbf{59.00} & \textbf{85.94} & \underline{92.20} & \multicolumn{1}{c|}{\textbf{79.05}} & \underline{46.20} & \underline{83.80} & 85.85 & \multicolumn{1}{c|}{71.95} & 40.10 & 72.45 & 77.02 & 63.19 \\ \midrule
\multicolumn{13}{c}{\textbf{LLM-basd Translators}} \\ \midrule
\multicolumn{1}{c|}{\textsc{\textbf{\cellcolor{lp!60}InstaTrans}}}                   & \cellcolor{lp!60}\underline{50.80} & \cellcolor{lp!60}\underline{81.23} & \cellcolor{lp!60}\underline{85.29} & \multicolumn{1}{c|}{\cellcolor{lp!60}\underline{72.44}} & \cellcolor{lp!60}\underline{45.60} & \cellcolor{lp!60}\underline{82.88} & \cellcolor{lp!60}\underline{85.65} & \multicolumn{1}{c|}{\cellcolor{lp!60}\underline{71.38}} & \cellcolor{lp!60}\underline{52.80} & \cellcolor{lp!60}\underline{87.69} & \cellcolor{lp!60}84.31 & \cellcolor{lp!60}\underline{74.93} \\
\multicolumn{1}{c|}{\textsc{\textbf{\cellcolor{lp!60}InstaTrans$_{pre}$}}}                   & \cellcolor{lp!60}\textbf{53.30} & \cellcolor{lp!60}\textbf{82.68} & \cellcolor{lp!60}\textbf{87.24} & \multicolumn{1}{c|}{\cellcolor{lp!60}\textbf{74.41}} & \cellcolor{lp!60}\textbf{46.10} & \cellcolor{lp!60}\textbf{83.04} & \cellcolor{lp!60}\textbf{86.45} & \multicolumn{1}{c|}{\cellcolor{lp!60}\textbf{71.86}} & \cellcolor{lp!60}\textbf{54.80} & \cellcolor{lp!60}\textbf{88.49} & \cellcolor{lp!60}\textbf{86.73} & \cellcolor{lp!60}\textbf{76.67} \\
\multicolumn{1}{c|}{\textbf{TowerBase}}                   & 38.80 & 79.01 & 76.88 & \multicolumn{1}{c|}{64.90} & 28.40 & 77.50 & 64.07 & \multicolumn{1}{c|}{56.66} & 19.80 & 83.39 & 61.22 & \multicolumn{1}{c}{54.80} \\
\multicolumn{1}{c|}{\textbf{TowerInstruct}}                   & 22.10 & 72.90 & 63.09 & \multicolumn{1}{c|}{52.70} & 16.20 & 70.88 & 56.73 & \multicolumn{1}{c|}{47.94} & 19.20 & 66.58 & 56.24 & 47.34  \\ \midrule
\multicolumn{13}{c}{\textbf{Commercial Translators}} \\ \midrule
\multicolumn{1}{c|}{\textbf{DeepL}}                   & \textbf{58.40} & \textbf{86.56} & \textbf{91.66} & \multicolumn{1}{c|}{\textbf{78.87}} & \textbf{47.60} & \textbf{85.47} & \underline{80.05} & \multicolumn{1}{c|}{\textbf{71.04}} & \textbf{55.40} & \textbf{89.34} & \textbf{82.21} & \textbf{75.65}  \\
\multicolumn{1}{c|}{\textbf{Google Translator}}                   & \underline{52.50} & \underline{83.39} & \underline{89.67} & \multicolumn{1}{c|}{\underline{75.19}} & \underline{43.80} & \underline{82.53} & \textbf{85.33} & \multicolumn{1}{c|}{\underline{70.55}} & \underline{51.10} & \underline{85.25} & \underline{82.08} & \underline{72.81}  \\
\multicolumn{1}{c|}{\textbf{Papago}}                   & 43.50 & 81.03 & 81.76 & \multicolumn{1}{c|}{68.76} & 39.20 & 81.76 & 81.92 & \multicolumn{1}{c|}{67.63} & 42.40 & 84.93 & 79.77 & 69.03  \\ \bottomrule
\end{tabular}
}
\caption{Machine translation performance comparison of \textsc{\textbf{InstaTrans}} against various state-of-the-art models. While \textsc{\textbf{InstaTrans}} is a further fine-tuned model based on the already fine-tuned version of SOLAR 10.7B, \textsc{\textbf{InstaTrans$_{pre}$}} is a fine-tuned model based on the pre-trained version of SOLAR 10.7B.}\label{tab:sota}
\end{table*}

\subsection{Experimental Settings}

\paragraph{Dataset and model details.}
In Step 1, we translated segments from OpenOrca~\cite{mukherjee2023orca} and Flan v2~\cite{longpre2023flan} into Korean.\footnote{Specifically, the segment from OpenOrca (respectively, Flan v2) comprises 2,000 (respectively, 3,000) randomly selected samples.} Then, in Step 2, given the outstanding performance of the {\it fine-tuned version} of SOLAR 10.7B~\cite{kim2023solar} as evidenced on the HuggingFace Open LLM Leaderboard~\cite{open-llm-leaderboard}, we selected this model as our base model for {\it further} fine-tuning with instruction datasets translated in Step 1, resulting in \textsc{\textbf{InstaTrans}}.

\vspace{1mm}
\paragraph{Competitors.}
We compare our \textsc{\textbf{InstaTrans}} with competitors in the following three categories:
\begin{itemize}[leftmargin=*, noitemsep, topsep=0.5pt] 
\item \textbf{Proprietary LLMs:} To validate the effectiveness of our prompt specifically designed to translate instruction datasets, we compare GPT-4 (our prompt) and GPT-3.5 (our prompt), which utilize our prompt along with function calling, with proprietary LLMs (GPT-4, GPT-3.5, and Gemini-Pro~\cite{anil2023gemini});
\item \textbf{LLM-based translators:} We compare \textsc{\textbf{InstaTrans}} with other LLM-based translators (TowerBase~\cite{towerbase} and TowerInstruct~\cite{towerinstruct}) that support the Korean language. Also, we present the performance of \textsc{\textbf{InstaTrans$_{pre}$}}, which is fine-tuned from the {\it pre-trained version} of SOLAR 10.7B, to verify the effectiveness of further fine-tuning strategy;
\item \textbf{Commercial translators:} To demonstrate the limitations of commercial translators (DeepL~\cite{deepl}, Google Translator~\cite{johnson2017google}, and Papago~\cite{lee2016papago}), we also present their performance.
\end{itemize}

\vspace{1mm}
\paragraph{Evaluation details.}
To evaluate the performance and quality of machine translation for instruction datasets, we used the sampled Ko-H3 datasets. These datasets were obtained for research purposes through a request to the Open Ko-LLM leaderboard~\cite{open-ko-llm-leaderboard}. They include the sampled Ko-arc, Ko-mmlu, and Ko-truthfulqa, which are composed of segments from the ARC~\cite{clark2018think}, MMLU~\cite{hendrycks2020measuring}, and TruthfulQA~\cite{lin2022truthfulqa} datasets translated into Korean, respectively.\footnote{Each dataset comprises 400 samples, and each sample is in the form of generation rather than multiple choice.} For the machine translation evaluation metrics, we utilize BLEU~\cite{papineni2002bleu}, COMET~\cite{rei2020comet}, and GEMBA~\cite{kocmi2023large}.

\subsection{Main Results}
We highlight the best results in bold and the second-best results with an underline in the following tables. Also, ours are shown in purple color.

\vspace{1mm}
\paragraph{RQ1: Comparison of machine translation performance.}
As shown in Table~\ref{tab:sota}, within the category of proprietary LLMs, GPT-4 (our prompt), GPT-4, and Gemini-Pro demonstrated a superior performance. In the realm of LLM-based translators, \textsc{\textbf{InstaTrans}} and \textsc{\textbf{InstaTrans$_{pre}$}} outperform TowerBase and TowerInstruct. In addition, TowerInstruct exhibits lower machine translation performance than TowerBase because it is fine-tuned to generate responses to any instruction, often providing answers without translation. Meanwhile, \textsc{\textbf{InstaTrans$_{pre}$}} outperforms \textsc{\textbf{InstaTrans}} in terms of machine translation performance.
Finally, DeepL achieved the state-of-the-art among commercial translators, even showing superior machine translation performance among all competitors.

\begin{table}[t]
\centering
\small
\resizebox{0.48\textwidth}{!}{
\begin{tabular}{c|ccc}
    \toprule 
    \textbf{Methods} & \textbf{Avg. C} & \textbf{Avg. I} & \textbf{Ratio (I / C) (\%)} \\ \midrule
    \multicolumn{1}{c|}{\textbf{\cellcolor{lp!60}GPT-4 (our prompt)}}     & \textbf{\cellcolor{lp!60}86.96} & \textbf{\cellcolor{lp!60}80.54} & \cellcolor{lp!60}92.62 \\
    \multicolumn{1}{c|}{\textbf{GPT-4}}     & 69.42  & 66.67 & \textbf{96.04} \\
    \multicolumn{1}{c|}{\textbf{{Gemini-Pro}}}     & 81.00 & 76.00 & \underline{93.83} \\
    \multicolumn{1}{c|}{\textsc{\textbf{\cellcolor{lp!60}InstaTrans}}}     & \underline{\cellcolor{lp!60}85.38} & \underline{\cellcolor{lp!60}78.25} & \cellcolor{lp!60}91.65 \\
    \multicolumn{1}{c|}{\textsc{\textbf{\cellcolor{lp!60}InstaTrans$_{pre}$}}}     & \cellcolor{lp!60}81.88 & \cellcolor{lp!60}75.71 & \cellcolor{lp!60}92.46 \\
    \multicolumn{1}{c|}{\textbf{DeepL}}     & 80.50 & 72.21 & 89.70 \\ \bottomrule
\end{tabular}
}
\caption{Quality comparison of instruction datasets translated by each translator.}
\label{tab:quality}
\end{table}

\vspace{1mm}
\paragraph{RQ2: Quality comparison of translated instruction datasets.}
Since machine translation metrics have limitations in assessing the quality of translated instruction datasets, we assessed the quality of instruction datasets translated by the top performers in each category using the same method outlined in Section~\ref{subsec:preliminary}.
Table~\ref{tab:quality} shows that i) the quality of instruction datasets translated by GPT-4 (our prompt) surpassed others in terms of both completeness and informativeness; ii) the quality of instruction datasets translated by \textsc{\textbf{InstaTrans}} was comparable to that of GPT-4 (our prompt); iii) LLM-based methods demonstrated a high ratio exceeding 90\%, in contrast to commercial translators, {\it i.e.}, Should they accomplish a complete translation, they successfully perform instruction-aware translation with a high ratio. Note that our \textsc{\textbf{InstaTrans}} successfully translates instruction datasets without requiring API costs, which means that LLMs can be extended to diverse languages at a relatively low cost.

\vspace{1mm}
\paragraph{RQ3: Effectiveness of translated instruction datasets.}
To validate the effectiveness of translated instruction datasets in fine-tuning LLMs {\it to align with user intentions}, we measure the performance of LLMs that have been fine-tuned with instruction datasets translated by the top performers in each category.
Specifically, we randomly sampled 5,000 examples from the Alpaca dataset and translated them using each translator. Then, we fine-tuned the pre-trained SOLAR 10.7B model with those translations.
To evaluate the performance of fine-tuned models, we utilize Ko MT-bench, an automated evaluation method based on GPT-4 that employs the Ko MT-bench dataset. This dataset is a translation of the MT-bench~\cite{zheng2023judging} by engaging experts in English-to-Korean translation.

\begin{figure}[t]
\centering
\begin{tikzpicture}
    \footnotesize
    \begin{axis}[
        ybar=0pt,
        width=8.0cm,
        height=3.0cm,
        bar width=0.45cm,
        bar shift=0pt,
        ylabel={Score},
        y label style={font=\tiny, at={(+0.08,0.5)}},
        ymin=5.5, ymax=7.0,
        symbolic x coords={\textsc{\textbf{IT}}, GPT-4\\(our prompt), \textsc{\textbf{IT$_{pre}$}}, Gemini-Pro, DeepL, GPT-4},
        xtick=data,
        ytick={5.5, 6.0, 6.5, 7.0},
        x tick label style={font=\tiny, align=center, text width=2cm, /pgf/number format/.cd,fixed,fixed zerofill,precision=2,/tikz/.cd},
        y tick label style={font=\tiny, /pgf/number format/.cd,fixed,fixed zerofill,precision=1,/tikz/.cd},
        ]
        \addplot [ybar, draw=purple!95, fill=purple!80] coordinates {
            (\textsc{\textbf{IT}}, 6.64375)
            (GPT-4\\(our prompt), 0)
            (\textsc{\textbf{IT$_{pre}$}}, 0)
            (Gemini-Pro, 0)
            (GPT-4, 0)
            (DeepL, 0)};
        \addplot [ybar, draw=purple!80, fill=purple!65] coordinates {
          (GPT-4\\(our prompt), 6.4625)};
        \addplot [ybar, draw=purple!65, fill=purple!50] coordinates {
          (\textsc{\textbf{IT$_{pre}$}}, 6.4)};
        \addplot [ybar, draw=purple!50, fill=purple!35] coordinates {
          (Gemini-Pro, 6.2)};
        \addplot [ybar, draw=purple!35, fill=purple!20] coordinates {
          (DeepL, 6.19375)};
      \addplot [ybar, draw=purple!20, fill=purple!5] coordinates {
          (GPT-4, 6.075)};
    \end{axis}
\end{tikzpicture}
\caption{Ko MT-bench score for models fine-tuned with instruction datasets translated by each translator. The x label represents the translators used for translation. For simplicity, we denote \textsc{\textbf{InstaTrans}} as \textsc{\textbf{IT}}.}
\label{fig:komt}
\end{figure}
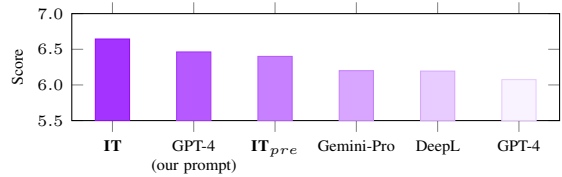

As illustrated in Figure~\ref{fig:komt}, the models fine-tuned with instruction datasets translated by \textsc{\textbf{InstaTrans}} achieved the highest score. This indicates that \textsc{\textbf{InstaTrans}} can generate high-quality non-English instruction datasets. These results also indicate that existing translation metrics do not sufficiently measure the performance of translation for instruction datasets.

\section{Conclusion}
In this paper, we have identified the successful translation of high-quality English instruction datasets as a promising direction for generating high-quality non-English instruction datasets, due to {\it tail phenomena}.
To successfully translate instruction datasets, we claim the critical importance of {\it completeness of translation} and {\it instruction-aware translation}.
In light of this, we proposed a new translation framework, named \textsc{\textbf{InstaTrans}} (INSTruction-Aware TRANSlation), which specializes in efficiently translating instruction datasets using LLMs.
We have empirically demonstrated that (1) \textsc{\textbf{InstaTrans}} successfully performs complete and instruction-aware translations, and (2) fine-tuning LLMs with datasets translated by \textsc{\textbf{InstaTrans}} can effectively improve their performance in the target language.

\section*{Acknowledgments}
This work was supported by Institute of Information \& Communications Technology Planning \& Evaluation(IITP) grant funded by the Korea government(MSIT) (No. RS-2024-00338140, Development of learning and utilization technology to reflect sustainability of generative language models and up-to-dateness over time).

\section*{Limitations}
In this study on the \textsc{\textbf{InstaTrans}} (INSTruction-Aware TRANSlation) framework, we acknowledge crucial limitations about the scope of research. Although we demonstrated the effectiveness of \textsc{\textbf{InstaTrans}}, our experiments were not conducted across a wide range of language pairs due to the absence of datasets to evaluate translation of instruction datasets. Generally, generating evaluation datasets necessitates substantial expenditure, as the engagement of human experts is essential. Due to these limitations, the scope of our evaluation was limited to English-to-Korean translations, which could potentially pose a challenge in ensuring the applicability of our findings across diverse languages.

Recently, numerous studies have been focusing on how to fine-tune the large language models (LLMs) for translation. However, due to the constraints of our computational resources, our study has not yet thoroughly investigated the methodology for fine-tuning to enhance their translation performance of instruction datasets. We proposed further fine-tuning LLMs that have already been fine-tuned, focusing soley on efficiency rather than the effectiveness in translating instruction datasets. Therefore, there is still room for improvement in the quality of the instruction dataset.

Despite these challenges, we demonstrated the critical importance of completeness of translation and instruction-aware translation in translating instruction datasets successfully. Furthermore, the \textsc{\textbf{InstaTrans}} framework marks a significant advancement in the field of translation instruction datasets, offering a scalable approach. Our research has established a foundation for future research in this critical area.

\section*{Ethics Statement}
In the development and evaluation of the \textsc{\textbf{InstaTrans}} (INSTruction-Aware TRANSlation) framework, as detailed in our study, we have adhered strictly to ethical guidelines and principles to ensure fairness and integrity throughout the research process. Our experimental setup was meticulously designed to provide an equitable and unbiased evaluation, taking into account the unique characteristics and challenges associated with translating instruction datasets for non-English languages. We have ensured that the translation process, facilitated by fine-tuned large language models (LLMs), was conducted with the utmost respect for the original content's integrity, aiming for completeness and instruction-awareness in translation to preserve the datasets' inherent attributes. We confirmed that all the data used in our experiments were free of licensing issues.

Furthermore, our approach to leveraging LLMs for enhancing translation quality was guided by ethical considerations, particularly in minimizing potential biases and ensuring the translations' fidelity to their source materials. The use of LLMs was conducted in a controlled and transparent manner, with careful examination of the models' outputs to detect and correct any errors or biases, thus maintaining the high-quality and reliability of the translated datasets.

In conclusion, our research on \textsc{\textbf{InstaTrans}} has been carried out with a strong commitment to ethical principles, ensuring that the experimental procedures were fair, the data handling was secure, and the translations produced were both accurate and sensitive to the nuances of instruction. We believe that our ethical approach not only underscores the validity of our research findings but also contributes positively to the broader field of language model development and application.

\bibliography{anthology, custom}

\clearpage
\appendix
\onecolumn

\section{Quality Comparison}
We provide a quantitative assessment for each task in Table~\ref{tab:qc_all} to deliver a multidimensional assessment of the quality of instruction datasets translated by various translators.

\begin{table*}[t!]
\centering
\footnotesize
{\renewcommand{\arraystretch}{1.25}
\resizebox{0.95\textwidth}{!}{
\begin{tabular}{c|ccccc|ccccc|c}
    \toprule 
    {} & \multicolumn{5}{c|}{\textbf{Completeness Score}} & \multicolumn{5}{c|}{\textbf{Informativeness Score}} & \textbf{Ratio (\%)} \\ \cline{2-11} 
    \textbf{Methods} & \textbf{Correction} & \textbf{Rephrase} & \textbf{Code} & \textbf{Others} & \textbf{Avg.} & \textbf{Correction} & \textbf{Rephrase} & \textbf{Code} & \textbf{Others} & \textbf{Avg.} & \textbf{(Avg. I / Avg. C)}\\ \midrule 
    \multicolumn{12}{c}{\textbf{Proprietary LLMs}} \\ \midrule
    \multicolumn{1}{c|}{\textbf{\cellcolor{lp!60}GPT-4 (our prompt)}}     & \cellcolor{lp!60}\textbf{87.67} & \cellcolor{lp!60}\textbf{89.83} & \cellcolor{lp!60}\underline{80.83} & \cellcolor{lp!60}\textbf{89.50} & \cellcolor{lp!60}\textbf{86.96} & \cellcolor{lp!60}\textbf{71.83} & \cellcolor{lp!60}\textbf{78.50} & \cellcolor{lp!60}\textbf{88.67} & \cellcolor{lp!60}\textbf{83.17} & \cellcolor{lp!60}\textbf{80.54} & \cellcolor{lp!60}92.62 \\
    \multicolumn{1}{c|}{\textbf{\cellcolor{lp!60}GPT-3.5 (our prompt)}}     & \cellcolor{lp!60}\underline{86.00} & \cellcolor{lp!60}78.83 & \cellcolor{lp!60}\textbf{81.00} & \cellcolor{lp!60}82.00 & \cellcolor{lp!60}\underline{81.96} & \cellcolor{lp!60}\underline{66.00} & \cellcolor{lp!60}68.50 & \cellcolor{lp!60}\underline{88.50} & \cellcolor{lp!60}78.83 & \cellcolor{lp!60}75.46 & \cellcolor{lp!60}92.07 \\
    \multicolumn{1}{c|}{\textbf{GPT-4}}     & 62.50 & 50.67 & \underline{80.83} & 83.67 & 69.42 & 50.83 & 50.00 & 88.33 & 77.50 & 66.67 & \underline{96.04} \\
    \multicolumn{1}{c|}{\textbf{GPT-3.5}}     & 48.83 & 55.00 & \underline{80.83} & 76.00 & 65.17 & 41.17 & 49.33 & 88.33 & 74.83 & 63.42 & \textbf{97.31} \\
    \multicolumn{1}{c|}{\textbf{Gemini-Pro}}     & 81.17 & \underline{85.83} & 68.33 & \underline{88.67} & 81.00 & 60.83 & \underline{74.33} & 88.00 & \underline{80.83} & \underline{76.00} & 93.83 \\ \midrule
    \multicolumn{12}{c}{\textbf{LLM-based Translators}} \\ \midrule
    \multicolumn{1}{c|}{\cellcolor{lp!60}\textsc{\textbf{InstaTrans$_{SFT}$}}}     & \cellcolor{lp!60}\textbf{83.00} & \cellcolor{lp!60}\textbf{84.83} & \cellcolor{lp!60}\textbf{84.17} & \cellcolor{lp!60}\underline{89.50} & \cellcolor{lp!60}\textbf{85.38} & \cellcolor{lp!60}\textbf{69.33} & \cellcolor{lp!60}\underline{72.67} & \cellcolor{lp!60}\textbf{88.67} & \cellcolor{lp!60}\underline{82.33} & \cellcolor{lp!60}\textbf{78.25} & \cellcolor{lp!60}91.65 \\
    \multicolumn{1}{c|}{\cellcolor{lp!60}\textsc{\textbf{InstaTrans$_{Base}$}}}     & \cellcolor{lp!60}\underline{78.50} & \cellcolor{lp!60}\underline{83.67} & \cellcolor{lp!60}\underline{75.17} & \cellcolor{lp!60}90.17 & \cellcolor{lp!60}\underline{81.88} & \cellcolor{lp!60}\underline{57.00} & \cellcolor{lp!60}\textbf{74.50} & \cellcolor{lp!60}\underline{88.33} & \cellcolor{lp!60}\textbf{83.00} & \cellcolor{lp!60}\underline{75.71} & \cellcolor{lp!60}92.46 \\
    \multicolumn{1}{c|}{\textbf{TowerBase}}     & 39.17 & 24.33 & 38.60 & 57.33 & 39.86 & 31.50 & 30.00 & 51.00 & 53.84 & 41.58 & \textbf{100.04}  \\
    \multicolumn{1}{c|}{\textbf{TowerInstruct}}     & 26.67 & 31.00 & 26.00 & 44.83 & 32.13 & 25.67 & 30.50 & 27.67 & 42.33 & 31.54 & \underline{98.16}  \\ \midrule
    \multicolumn{12}{c}{\textbf{Commercial Translators}} \\ \midrule
    \multicolumn{1}{c|}{\textbf{DeepL}}     & 73.50 & \underline{81.00} & \underline{79.17} & \underline{88.33} & \underline{80.50} & \textbf{55.67} & \underline{71.00} & \textbf{79.67} & \textbf{82.50} & \textbf{72.21} & \textbf{89.70} \\
    \multicolumn{1}{c|}{\textbf{Google Translator}}     & \underline{75.00} & \textbf{82.00} & \textbf{82.83} & \textbf{89.00} & \textbf{82.21} & 51.33 & \textbf{72.00} & \underline{79.00} & \underline{82.00} & \underline{71.08} & 86.46 \\
    \multicolumn{1}{c|}{\textbf{Papago}}     & \textbf{75.50} & 78.00 & 67.50 & 87.83 & 77.21 & \underline{53.33} & 67.33 & 65.50 & 81.00 & 66.79 & \underline{86.50} \\ \bottomrule
\end{tabular}
}
}
\caption{Quality comparison of instruction datasets translated by various translators.}\label{tab:qc_all}
\end{table*}

\end{document}